%% file: main.tex
\documentclass[10pt,twocolumn,letterpaper]{article}

\usepackage{iccv}              
\usepackage{graphicx}
\usepackage{amsmath}
\usepackage{amssymb}
\usepackage{multirow}
\usepackage{booktabs}
\usepackage{colortbl}  
\usepackage{xcolor}
\usepackage{array}  
\usepackage{booktabs}
\usepackage[accsupp]{axessibility}


\definecolor{iccvblue}{rgb}{0.21,0.49,0.74}
\usepackage[pagebackref,breaklinks,colorlinks,allcolors=iccvblue]{hyperref}

\def\paperID{12970} 
\def\confName{ICCV}
\def\confYear{2025}

\title{Adversarial Reconstruction Feedback for Robust Fine-grained Generalization}

\author{
Shijie Wang$^1$, Jian Shi$^2$, Haojie Li$^{1}$\thanks{Corresponding author: hjli@sdust.edu.cn } \\
   \textsuperscript{\rm 1}College of Computer and Engineering, Shandong University of Science and Technology, China \\
\textsuperscript{\rm 2} School of Software, Dalian University of Technology, China \\}
\begin{document}
\maketitle
\begin{abstract} 
Existing fine-grained image retrieval (FGIR) methods predominantly rely on supervision from predefined categories to learn discriminative representations for retrieving fine-grained objects. 
However, they inadvertently introduce category-specific semantics into the retrieval representation, creating semantic dependencies on predefined classes that critically hinder generalization to unseen categories.
To tackle this, we propose AdvRF, a novel adversarial reconstruction feedback framework aimed at learning category-agnostic discrepancy representations. 
Specifically, AdvRF reformulates FGIR as a visual discrepancy reconstruction task via synergizing category-aware discrepancy localization from retrieval models with category-agnostic feature learning from reconstruction models.
The reconstruction model exposes residual discrepancies overlooked by the retrieval model, forcing it to improve localization accuracy, while the refined signals from the retrieval model guide the reconstruction model to improve its reconstruction ability.
Consequently, the retrieval model localizes visual differences, while the reconstruction model encodes these differences into category-agnostic representations. This representation is then transferred to the retrieval model through knowledge distillation for efficient deployment.
Quantitative and qualitative evaluations demonstrate that our AdvRF achieves impressive performance on both widely-used fine-grained and coarse-grained datasets.
\end{abstract}

\section{Introduction}
\label{sec:intro}

\begin{figure}
     \centering
     \begin{subfigure}[b]{1\linewidth}
         \centering
         \includegraphics[width=\linewidth]{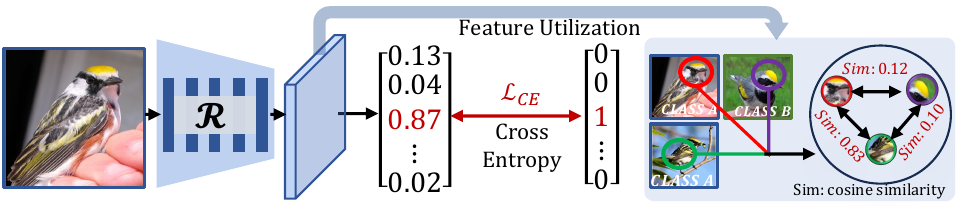}
         \caption{Classification-based Representation Learning}
         \label{introd1}
     \end{subfigure}

     \begin{subfigure}[b]{1\linewidth}
         \centering
         \includegraphics[width=\linewidth]{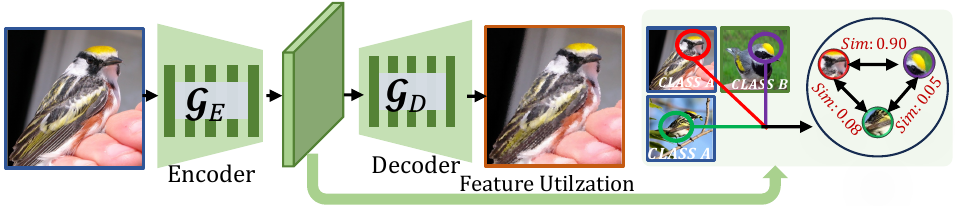}
         \caption{Reconstruction-based Representation Learning}
         \label{introd2}
         
     \end{subfigure}
             \caption{
             Motivation of the proposed AdvRF. 
             (a) Classification-based  representation learning easily embeds predefined category semantics into representations, causing visually similar cues across subcategories to appear dissimilar in feature space while visually dissimilar cues within the same subcategory appear similar.
             (b) Reconstruction-based representation learning allows for a focus on object details and the contextual semantics of both the object and its parts, based on their appearance, thereby capturing category-agnostic representations. This facilitates a deeper understanding of unseen categories by enabling precise appearance modeling using category-agnostic visual descriptions.
        }
       
\end{figure}

Fine-grained image retrieval (FGIR) aims to retrieve visually similar subcategories, even those that were unseen during the training phase.
It plays a vital role in numerous vision applications from fashion industry, \textit{e.g.}, retrieval of diverse types of clothes~\cite{DBLP:conf/cvpr/LiuLQWT16,DBLP:conf/cvpr/AkKLT18, DBLP:conf/aaai/WangWLDL20, DBLP:journals/tcsv/LiLPWYW24}, to environmental conservation, \textit{e.g.}, retrieving endangered species~\cite{DBLP:conf/cvpr/ElhoseinyZZE17, 10091192, DBLP:journals/ijcv/WangWLCOT24, DBLP:conf/cvpr/WangWYLLL20}. 
Given its significance, a substantial body of research has focused on learning discriminative and generalizable embeddings to enhance the performance of FGIR.

Current work on FGIR tasks~\cite{DBLP:conf/cvpr/LimYP022,DBLP:conf/wacv/MoskvyakMDB21,DBLP:conf/cvpr/RothVA22,DBLP:conf/eccv/LiuHZYTW22, DBLP:conf/nips/WangCL0O023, DBLP:journals/pami/WeiSSWP23} has achieved promising results by introducing metric constraints or designing localization schemes to capture diverse visual discrepancies from visually similar objects.
However, these approaches fundamentally couple discrepancy modeling with predefined category supervision, inadvertently embedding category-specific semantics into the retrieval representations. 
As illustrated in Fig.~\ref{introd1}, this coupling leads to two paradoxical phenomena: (1) visually similar bird heads from different subcategories are represented differently due to category divergence, whereas (2) dissimilar bird heads and wings within the same subcategory are clustered due to sharing the same category.
Consequently, the model’s representation power relies heavily on predefined category semantics, struggling to interpret unseen subcategories based on actual visual appearances.

Fortunately, object reconstruction tasks inherently focus on pixel-level fidelity and contextual coherence, requiring the model to preserve appearance details and their contextual semantics without embedding category-specific semantics \cite{DBLP:journals/pami/HanLB21, DBLP:conf/3dim/JiangJGZ24}. 
As demonstrated in Fig.~\ref{introd2}, this property enables the reconstruction model's encoder to generate similar feature representations for visually similar parts across different subcategories.
Building on this insight, we consider to leverage the inherent category-agnostic representation capability of object reconstruction models to eliminate dependencies on predefined category semantics in visual discrepancy modeling.
However, naively training a fine-grained reconstruction model proves inadequate for FGIR, as it focuses on modeling the entire image appearance, including irrelevant background information, instead of emphasizing key visual discrepancies.
Therefore, one natural question arises: is it possible to synergize category-aware discrepancy localization from retrieval models with category-agnostic feature learning from reconstruction models to model visual discrepancies using category-agnostic representations?

To answer this, we propose AdvRF, a novel Adversarial Reconstruction Feedback framework that reformulates FGIR as a visual discrepancy reconstruction task. AdvRF synergizes a retrieval model and a reconstruction model within an adversarial learning paradigm inspired by Generative Adversarial Networks (GANs) \cite{DBLP:journals/cacm/GoodfellowPMXWO20}. The framework operates through alternating optimization: while the retrieval model pinpoints subtle discrepancies within objects, the reconstruction model represents these discrepancies with category-agnostic representations, creating a self-reinforcing cycle where each component iteratively challenges and reinforces each other. Specifically, the reconstruction model exposes residual discrepancies overlooked by the retrieval model, forcing it to improve localization accuracy, while the refined signals from the retrieval model guide the reconstruction model to improve its reconstruction ability. This adversarial interplay progressively achieves precise localization of visual discrepancies, while eliminating predefined category semantics in discrepancy modeling. For efficient deployment, AdvRF distills category-agnostic discrepancy representations purified by the reconstruction model into the retrieval model through lightweight knowledge distillation.

Our main contribution are summarized below:

\begin{itemize}
    \item To the best of our knowledge, we are the first to reformulate FGIR tasks as a visual discrepancy reconstruction process, improving its generalization.

    \item An adversarial reconstruction feedback framework, \textit{i.e.}, AdvRF, is proposed. AdvRF establishes an adversarial pipeline to alternately train the retrieval and reconstruction models, effectively capturing category-agnostic discrepancies essential for representing unseen categories.

    \item Extensive experiments show that our AdvRF achieves state-of-the-art performance on widely-used fine-grained and coarse-grained retrieval benchmarks.
    
\end{itemize}

\begin{figure*}[t]
\begin{center}
   \includegraphics[width=1\linewidth]{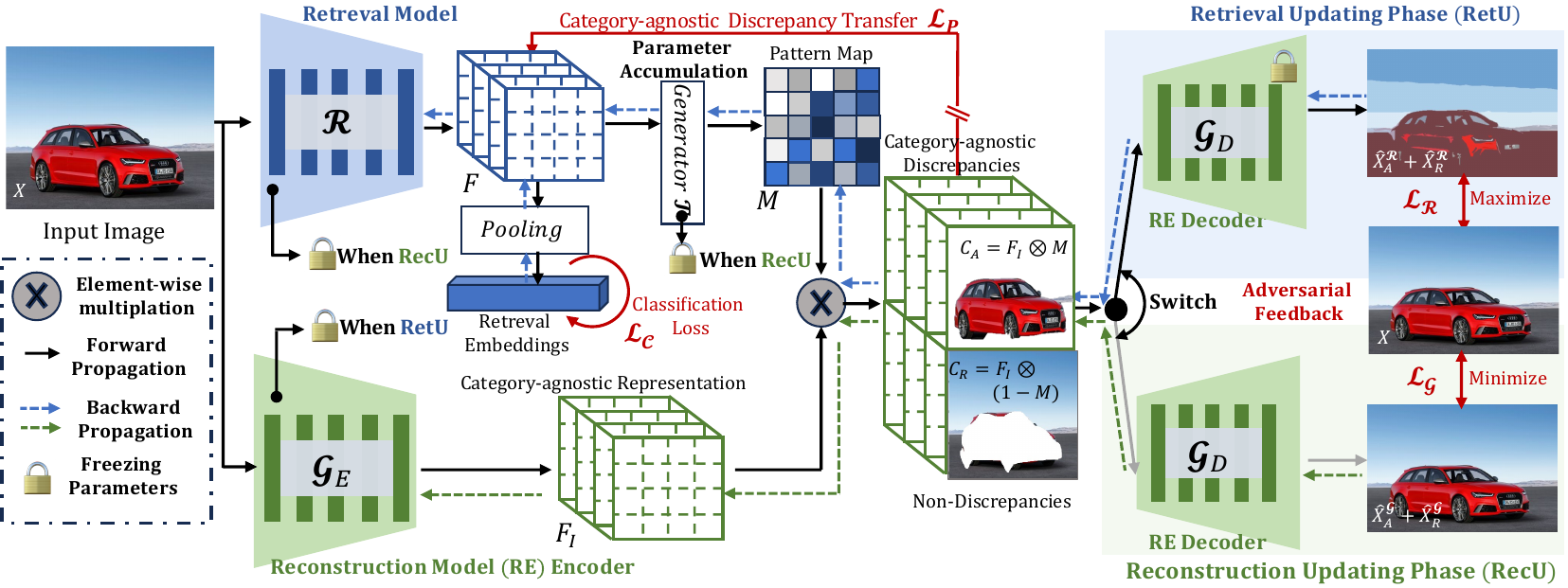}
\end{center}
\setlength{\abovecaptionskip}{-0.1cm} 
   \caption{Detailed illustration of {
   \bf adversarial reconstruction feedback}. See \S\ref{adv} for more details. }
\label{approach}
\end{figure*}

\section{Related Work}
\textbf{Fine-grained image retrieval} can be broadly categorized into two groups. The first group, \textit{localization-based scheme}, focuses on localizing object or its details to facilitate the retrieval of visually similar objects
\cite{DBLP:journals/tip/WeiLWZ17,DBLP:conf/ijcai/ZhengJSWHY18,DBLP:conf/wacv/MoskvyakMDB21,DBLP:conf/aaai/WangWLO22}. 
CaRA \cite{DBLP:journals/pami/WangCWLOT24} implements a rectified activation strategy to enhance the localization of object details. 
The second group, \textit{metric-based schemes}, seeks to learn an embedding space where similar examples are closely aligned, while dissimilar ones are pushed apart~\cite{DBLP:conf/iccv/KoGK21,DBLP:conf/icml/RothMOCG21,DBLP:conf/iccv/ZhengZL021,DBLP:conf/cvpr/KimKCK21,DBLP:conf/cvpr/ZhengWL021}. 
NIA \cite{DBLP:conf/cvpr/RothVA22} enforces unique translatability of samples from their respective class proxies to bring the distance of samples with the same subcategory closer.
Unlike these approaches, PLEor \cite{DBLP:conf/cvpr/WangCLWO023} incorporates category-specific language descriptions based on the CLIP model to guide the model in representing visual discrepancies.
However, such methods, which rely on representing similar objects through predefined categories, unintentionally embed category-specific semantic information into the retrieval representations, thereby limiting their ability to generalize to unseen categories.
To address this issue, we propose AdvRF, which incorporates an adversarial pipeline to acquire category-agnostic discrepancies derived from the reconstructed features  of objects.

\textbf{Adversarial learning} has gained extensive application across multiple domains, including generative adversarial networks (GANs) \cite{DBLP:conf/cvpr/FeiFZHWW23, DBLP:conf/cvpr/WangZM22, DBLP:conf/cvpr/ChenXTZHG21}, person re-identification \cite{DBLP:conf/cvpr/ZhengYY00K19}, and domain adaptation \cite{DBLP:conf/icml/GaninL15, DBLP:conf/cvpr/HuKSC18}. 
The essence of adversarial learning lies in minimizing distributional discrepancies between target and source domains by counteracting adversarial attacks.  
Furthermore, existing work such as OpenGAN \cite{DBLP:conf/iccv/KongR21} seeks to enhance the model's generalization to unseen categories by training a robust open-vs-closed discriminator that distinguishes between synthesized fake data and real data. 
In contrast to OpenGAN, which improves generalization through the generation of fake samples, AdvRF enhances generalization by capturing class-agnostic discrepancies via adversarial learning between retrieval and reconstruction models from a feature perspective.



\section{Adversarial Reconstruction Feedback }
\label{adv}
The core of AdvRF, as depicted in Fig.~\ref{approach}, lies in its innovative collaboration between a retrieval model and a reconstruction model. The retrieval model pinpoints discrepancy locations within objects, while the reconstruction model, leveraging these locations, generates category-agnostic representations of these discrepancies. This is achieved through adversarial feedback learning, where the two models create a self-reinforcing cycle where each component iteratively challenges and reinforces each other. Furthermore, these discrepancies are back into the retrieval model via knowledge distillation, significantly boosting computational efficiency during test.

\subsection{Network Architecture}
\textbf{Fine-grained Retrieval Model $\mathcal{R}$.}
It is designed to extract robust object representations and generate the final retrieval embeddings. Given an input image $\mathbf{X}$, we denote $\mathbf{F} \in \mathbb{R}^{C\times H\times W}$ as the $C$-dimensional feature tensor with $H \times W$ spatial dimensions, encoded by a backbone network $\mathbf{F}=\mathcal{R}(\mathbf{X})$.
Traditionally, the most prevalent approach for fine-grained retrieval tasks embeds the full feature tensor $\mathbf{F}$ through global average pooling (GAP, $g(\cdot)$), which computes the mean across the $H \times W$ spatial planes.
This process yields the final retrieval embeddings $\mathbf{E_R} \in \mathbb{R}^C = g(\mathbf{F})$. 
Importantly, our proposed AdvRF framework maintains computational efficiency, as only the retrieval model is needed at inference time. 

\textbf{Fine-grained Reconstruction Model $\mathcal{G}$.}
It consists of an encoder $\mathcal{G}_E$ and a decoder $\mathcal{G}_D$, with the detailed architecture outlined below.
For the encoder $\mathcal{G}_E$, we use a lightweight network, specifically ResNet34, with pooling layers omitted from the last two blocks in our experimental setup. To emphasize subtle discrepancies within fine-grained objects, the encoder aggregates feature maps from the last three blocks into a final representation, enabling efficient encoding of input signals. This aggregation process applies $1 \times 1$ convolutional layers, which distills subtle details from low-level features while capturing semantic information from high-level representations.

The decoder, $\mathcal{G}_D$, adopts a U-Net architecture \cite{DBLP:conf/miccai/RonnebergerFB15} with eight downsampling blocks, seven upsampling blocks, and a final colorization block to ensure high-fidelity reconstruction of the inputs. 
Each downsampling block consists of a $4 \times 4$ convolutional layer with stride 2, followed by a normalization layer and LeakyReLU activation. Similarly, each upsampling block contains a transposed convolutional layer with stride 2, followed by a normalization layer and LeakyReLU activation.


\subsection{Category-agnostic Discrepancy Acquisition}
Considering that the reconstruction model is primarily designed to recover pixel-level details from inputs, it faces difficulties in identifying which visual cues represent meaningful discrepancies for identifying visually similar objects. 
To address this, we design a category-agnostic discrepancy acquisition module that captures visual discrepancies through category-agnostic representations produced by the reconstruction model.

\textbf{Discrepancy Decoupling.} 
Since the retrieval model is designed to capture visual discrepancies within images, we leverage its representation as a foundation to identify and localize both visual discrepancy and non-discrepancy regions.
Therefore, we map the retrieval representation \(\mathbf{F}\) into a pattern map \(\mathbf{\hat{M}} \in \mathbb{R}^{H \times W}\), which serves to indicate the locations of discrepancies. $\mathbf{\hat{M}}$ can be generated by a light-weight generator $\mathcal{T}(\cdot)$ as below:
\begin{equation}
   \mathbf{\hat{M}} = \sigma(\mathcal{T}(\mathbf{F})) ,
   \label{eq1}
\end{equation}
where $\sigma(\cdot)$ is the sigmoid activation function, $\mathcal{T}(\cdot)$ is a convolutional layer with kernel size 1. 

However, optimizing the pattern map via back-propagation of the loss function primarily tunes the parameters of the lightweight generator, with minimal impact on the retrieval model’s parameters. 
Consequently, the output of the retrieval model may still include non-discriminative information, potentially impairing retrieval performance. 
To ensure the retrieval model is optimized to focus exclusively on visual discrepancies, we introduce an mean generator with the same architecture to produce a refined pattern map.
 In this way, Eq.~\ref{eq1} can be rewritten as:
\begin{equation}
    \mathbf{\hat{M}} = \sigma(\mathrm{E}(\mathcal{T})(\mathbf{F})),
\end{equation}
where $\mathrm{E}(\mathcal{T})$ denotes the mean generator without learnable parameters. 
Its parameters can be updated in a temporal average manner. Concretely, at the $t$-th iteration, parameters $\mathrm{E}(\mathcal{T})$ are accumulated by:
\begin{equation}
    \mathrm{E}^{(t)}(\mathcal{T})[\theta] = (1-\delta) \cdot \mathrm{E}^{(t-1)}(\mathcal{T})[\theta] + \delta \cdot \theta,
    \label{eq3}
\end{equation}
where $\mathrm{E}^{(t)}(\mathcal{T})[\theta]$ and $\mathrm{E}^{(t-1)}(\mathcal{T})[\theta]$ denote the parameters of the mean generator in current iteration and last iteration, respectively. The mean generator is initialized as $\mathrm{E}^{(0)}(\mathcal{T})[\theta] = \theta$. The hyper-parameter $\delta$ is the updating ratio within the range of $(0,1]$. 

\textbf{Category-agnostic Discrepancy Transfer.}
We feed the input image $\mathbf{X}$ into the encoder $\mathcal{G}_E$ of the reconstruction model to obtain the category-agnostic representation $\mathbf{\hat{F}_I} = \mathcal{G}_E(\mathbf{X}) \in \mathbb{R}^{C \times H \times W}$, and then resize both the pattern map $\mathbf{\hat{M}}$ and $\mathbf{\hat{F}_I}$ to match the size of the original input images, ensuring high fidelity during the subsequent reconstruction process. Using the amplified pattern map $\mathbf{M}$, we decompose the amplified image representation $\mathbf{F_I}$ into the category-agnostic visual discrepancies $\mathbf{C_A}$ and the non-discrepancy representation $\mathbf{C_R}$ as follows:
\begin{equation}
    \mathbf{C_A} = \mathbf{F_I} \odot \mathbf{M}, \quad  \mathbf{C_R} = \mathbf{F_I} \odot (1 - \mathbf{M}).
    \label{eq4}
\end{equation}
Here, $\odot$ denotes element-wise multiplication. 

Considering that employing both models concurrently is time-consuming and memory-intensive for retrieval evaluation, we design a category-agnostic discrepancy parameterization constraint. 
Formally, the category-agnostic discrepancy serves as the supervisory signal and eliminates the need for gradient updates to adjust retrieval representations:
\begin{equation}
    \mathcal{L}_\mathcal{P} = ||\mathbf{E_R}-g(\mathbf{C_A})||,
\end{equation}
where $||\cdot||$ refers to the Frobenius norm.
This constraint directly optimizes the parameters within the retrieval model, ensuring that its output representations exclusively capture category-agnostic visual discrepancies while eliminating category-specific semantics. As a result, the retrieval model gains the ability to pinpoint object discrepancies and characterize them based solely on their visual appearance, even when encountering unseen categories. Importantly, the contextual semantics of the discrepancies are still preserved, as the reconstruction process inherently considers the contextual semantics of both the object and its parts.


\subsection{Adversarial Feedback Learning} 
To acquire category-agnostic discrepancy representations, we introduce an adversarial feedback learning strategy inspired by GANs, which synergizes a retrieval model with a reconstruction model.
Through adversarial feedback interplay, the reconstruction model exploits residual discrepancies overlooked by the retrieval model for reconstruction, thereby challenging the retrieval model and enhancing object recovery. 
Concurrently, the retrieval model dynamically refines its discrepancy localization based on feedback from the reconstruction model, further challenging the reconstruction model and improving discrepancy localization.

\textbf{Reconstruction Feedback.}
Given the category-agnostic discrepancies $\mathbf{C_A}$ and the non-discrepancy representation $\mathbf{C_R}$, we input them separately into the decoder $\mathcal{G}_D(\cdot)$ of the reconstruction model to obtain the reconstructed regions:
\begin{equation}
    \mathbf{X_A^\mathcal{G}} = \mathcal{G}_D(\mathbf{C_A}), \quad 
    \mathbf{X_R^\mathcal{G}} = \mathcal{G}_D(\mathbf{C_R}).
    \label{eq6}
\end{equation}
Here, $\mathbf{X_A^\mathcal{G}}$ and $\mathbf{X_R^\mathcal{G}}$ represent the reconstructed images, respectively.
Considering that the reconstruction model utilizes the residual discrepancies overlooked by the retrieval model to reconstruct discrepancies, we should evaluate the quality of discrepancy reconstruction using the image generated by non-discrepancy representation \(\mathbf{C_R}\). Therefore, we minimize the difference between the reconstructed image $\mathbf{X_R^\mathcal{G}}$ and the input image $\mathbf{X}$, within the discrepancy region:
\begin{equation}
    \mathcal{L}^{R}_{\mathcal{G}} = || \mathbf{M} \odot (\mathbf{X}-\mathbf{X_R^\mathcal{G}})||.
    \label{eq7}
\end{equation}
Here, $\mathbf{M}$ acts as a spatial attention map to localize discrepancies, maintaining intrinsic invariance across feature/pixel spaces. It can be regarded as a soft validation mask to direct the reconstruction model's focus to critical discrepancy regions during regeneration.

Similarly, the reconstruction model also leverages the residual non-discrepancy information from the category-agnostic representation \(\mathbf{C_A}\) to reconstruct the non-discrepancy regions. Hence, we also impose another reconstruction constraint for minimizing their differences between the reconstructed image $\mathbf{X_A^\mathcal{G}}$ and the original images $\mathbf{X}$:
\begin{equation}
    \mathcal{L}^{A}_{\mathcal{G}} = || (1 - \mathbf{M}) \odot (\mathbf{X}-\mathbf{X_A^\mathcal{G}})||,
    \label{eq8}
\end{equation}
where $(1 - \mathbf{M})$  indicates the non-discrepancy region.

Therefore, the total loss for training the reconstruction model could be integrated as:
\begin{equation}
    \mathcal{L}_\mathcal{G}^{RecF} = \mathcal{L}^{A}_{\mathcal{G}} + \mathcal{L}^{R}_{\mathcal{G}}.
\end{equation}
To prevent the retrieval model from optimizing towards less accurate discrepancy localization, we freeze its parameters and exclusively back-propagate the loss gradients to the reconstruction model, including both its encoder and decoder. 


\textbf{Retrieval Feedback.}
Unlike the reconstruction feedback, which aims to improve the reconstruction ability of the reconstruction model, the retrieval feedback focuses on precisely locating discrepancies within objects. 
It leverages the reconstruction model as an evaluator to assess the accuracy of discrepancy localization.
The above process is much like how the generator receives feedback from the discriminator in GANs.
Formally, we also need to feed $\mathbf{C_A}$ and $\mathbf{C_R}$ into $\mathcal{G}_D $ to obtain the reconstructed images as below:
\begin{equation}
    \mathbf{\hat{X}_A^{\mathcal{R}}} = \mathcal{G}_D(\mathbf{C_A}), \quad 
    \mathbf{\hat{X}_R^{\mathcal{R}}} = \mathcal{G}_D(\mathbf{C_R}),
\end{equation}
where $\mathbf{\hat{X}_A^{\mathcal{R}}}$ and $\mathbf{\hat{X}_R^{\mathcal{R}}}$ represent the reconstructed images, respectively.

This optimization strategy is similar to optimizing the generator based on feedback from the discriminator.
In other words, the discrepancy localization provided by the retrieval model makes it challenging for the reconstruction model to accurately reconstruct the object.
Therefore, given the reconstructed images $\mathbf{\hat{X}_R^{\mathcal{R}}}$ and the original image $\mathbf{X}$ on the discrepancy regions, we impose a reconstruction constraint to maximize their differences:
\begin{equation}
    \mathcal{L}^R_\mathcal{R} = -|| \mathbf{M} \odot (\mathbf{X}-\mathbf{\hat{X}_R^\mathcal{R}})||.
\end{equation}
Importantly, the minus sign means that minimizing the loss leads the framework to maximize the difference between the $\mathbf{X}$ and $\mathbf{\hat{X}_R^\mathcal{R}}$.
Similarly, we impose an additional reconstruction constraint to maximize the differences between the reconstructed and original non-discrepancy regions:
\begin{equation}
    \mathcal{L}^A_\mathcal{R} = -||(1- \mathbf{M}) \odot (\mathbf{X}-\mathbf{\hat{X}_A^{\mathcal{R}}})||.
\end{equation}
Finally, the total loss for training the retrieval model to produce the accurate pattern map is:
\begin{equation}
    \mathcal{L}_\mathcal{R}^{RetF} = \mathcal{L}^A_\mathcal{R} + \mathcal{L}^R_\mathcal{R}.
\end{equation}
Similarly, to prevent compromising the reconstruction capabilities of the reconstruction model, we freeze its parameters and exclusively backpropagate the loss gradients to the retrieval model.

\subsection{Alternating Training Strategy}
AdvRF implements an iterative alternating protocol where the reconstruction and retrieval models cyclically enhance each other's improvement. In each training epoch, the following steps are performed:
\begin{enumerate}
\item \textbf{Reconstruction Updating Phase (RecU)}: The parameters of the retrieval model \(\Theta_\mathcal{R}\) are frozen, and the reconstruction model \(\Theta_\mathcal{G}\) is updated using the loss function:
\begin{equation}
\mathcal{L}_\mathcal{G}^{RecU}[\Theta_\mathcal{G}] = \alpha \cdot \mathcal{L}_\mathcal{G}^{RecF}.
\label{eq14}
\end{equation}
 \item \textbf{Retrieval Updating Phase (RetU)}: Freezing $\Theta_\mathcal{G}$, we then refine $\Theta_\mathcal{R}$ using multi-task learning with:
\begin{equation}
\mathcal{L}_\mathcal{R}^{RetU} [\Theta_\mathcal{R}] = \mathcal{L}_\mathcal{C} + \beta\cdot \mathcal{L}_\mathcal{P} + \gamma\cdot \mathcal{L}_\mathcal{R}^{RetF}.
\label{eq15}
\end{equation}
\end{enumerate}
This alternating training strategy repeats until joint convergence, with $\alpha$, $\beta$ and $\gamma$ dynamically balancing task-specific gradients.

\begin{table}\centering
	\caption{ Comparison of performance and efficiency on CUB-200-2011 using different combinations of constraints. The first row indicates that we use classification-based feedback as supervision, to replace the proposed AdvRF for comparison. "T" is the time of extracted retrieval embeddings.
	}
	\begin{tabular}{cccccccc}
	
		\toprule[1pt]
  \multicolumn{2}{c}{$\mathcal{L}_{\mathcal{G}}^{RecF}$}& &\multicolumn{2}{c}{$\mathcal{L}_{\mathcal{R}}^{RetF}$}&\multirow{2}{*}{$\mathcal{L}_{\mathcal{P}}$} & \multicolumn{2}{c}{Performance} \\
  \cline{1-2} \cline{4-5} \cline{7-8} 
		 $\mathcal{L}_{\mathcal{G}}^A$&$\mathcal{L}_{\mathcal{G}}^R$&&$\mathcal{L}_\mathcal{R}^{A}$&$\mathcal{L}_\mathcal{R}^{R}$&&R@1&T\\
		\toprule[0.7pt]
		&&&&&&66.3\% &21.1ms\\
		\hline
		\checkmark&&&\checkmark&&&73.7\% &36.7ms\\
		&\checkmark&&&\checkmark&&73.4\%&36.7ms \\
		\checkmark&\checkmark&&\checkmark&\checkmark&&76.8\%&36.7ms \\
		\rowcolor{gray!30}
		\checkmark&\checkmark&&\checkmark&\checkmark&\checkmark&{\bf 76.6\%}&{\bf 21.1ms} \\
		\hline
		
		\hline
	\end{tabular}\\
	\label{t1}
\end{table}

\begin{table}\centering
	\caption{ Evaluation results of retrieval performance on CUB-200-2011 dataset with diverse feedback.
	}
 \setlength{\tabcolsep}{14pt}
	\begin{tabular}{ll}
		\toprule[1pt]
		Feedback Type & Recall@1\\
		\toprule[0.7pt]
		Classification-based Feedback& 66.3\%\\
		Reconstruction-based Feedback& 62.2\%\\
        \hline 
		 \rowcolor{gray!30}
		Our AdvRF & {\bf76.6\%}\\
		\toprule[1pt]
		
	\end{tabular}\\
	\label{t2}
\end{table}

\begin{table*}[!t]\centering
	\caption{ Compared with competitive methods on CUB-200-2011, Stanford Cars 196 and FGVC Aircraft datasets. "Arch" represents the architecture of utilizing backbone network. "R50" denotes Resnet50 \cite{He2015Deep} backbone network. 
} 
	\begin{tabular}{l|c||cccc|cccc|cccc}
		\toprule[1pt]
		
       \multirow{2.5}{*}{Method} & \multirow{2.5}{*}{Arch}& \multicolumn{4}{c|}{CUB-200-2011} &\multicolumn{4}{c|}{Stanford Cars 196} & \multicolumn{4}{c}{FGVC Aircraft}\\
		\cline{3-14}
		 &  & 1&2&4&8& 1&2&4&8& 1&2&4&8 \\
			\toprule[0.7pt]
	 	SCDA $_{\rm  TIP{17}}$ \cite{DBLP:journals/tip/WeiLWZ17}& R50& 57.3 & 70.2 & 81.0 &88.4& 48.3& 60.2&71.8&81.8&  56.5& 67.7&77.6&85.7\\

CRL $_{\rm  IJCAI{18}}$ \cite{DBLP:conf/ijcai/ZhengJSWHY18} & R50&  62.5 & 74.2 & 82.9 &89.7& 57.8& 69.1&78.6&86.6& 61.1& 71.6&80.9&88.2\\

HDCL $_{\rm  IJON{21}}$ \cite{DBLP:journals/ijon/ZengLWZCL21} & R50&  69.5 & 79.6 & 86.8 &92.4& 84.4& 90.1&94.1&96.5& 71.1& 81.0&88.3&93.3\\
CEP $_{\rm  ECCV{20}}$ \cite{DBLP:conf/eccv/BoudiafRZGPPA20}& R50&69.2 & 79.2 & 86.9 &91.6&  89.3& 93.9&96.6&98.1& -& -& -&-\\
CaRA $_{\rm  TPAMI{24}}$ \cite{DBLP:journals/pami/WangCWLOT24} & R50& 73.9 & 82.2 & 89.4 &93.6& 94.1& 96.9&98.2&98.9&84.3& 90.4&94.2&96.3\\
FRPT $_{\rm  AAAI{23}}$ \cite{DBLP:conf/aaai/WangCWLO023} & R50& 74.3 &83.7& 89.8& 94.3 &91.1 &95.1 &97.3& 98.6 &77.6 &85.7& 91.4 &95.6\\
	\toprule[0.7pt]
  	DGCRL $_{\rm  AAAI{19}}$ \cite{DBLP:conf/aaai/ZhengJSZWH19}& R50&  67.9 & 79.1 & 86.2 &91.8& 75.9& 83.9&89.7&94.0& 70.1& 79.6&88.0&93.0\\

		DAS $_{\rm  ECCV{22}}$ \cite{DBLP:conf/eccv/LiuHZYTW22} & R50 &  69.2&79.3&87.1&92.6&87.8&93.2&96.0&97.9&- &- &- &-\\
		CBML $_{\rm  TPAMI{23}}$ \cite{DBLP:journals/pami/KanHCLMH23}& R50& 69.9 &80.4 &87.2 &92.5& 88.1 & 92.6 &95.4&97.4 &-& -& -&-\\                                    

 	NIR $_{\rm  CVPR{22}}$ \cite{DBLP:conf/cvpr/RothVA22} & R50& 70.5 & 80.6 & -&-&89.1 & 93.4 & -&-&-&-& -& - \\
 

		HIST $_{\rm  CVPR{22}}$ \cite{DBLP:conf/cvpr/LimYP022} & R50& 71.4 &  81.1& 88.1&-&89.6& 93.9& 96.4&-&-&-& -& - \\

 IDML $_{\rm  TPAMI{24}}$ \cite{DBLP:journals/pami/WangZZZL24} & R50 & 70.7& 80.2& - & -&90.6& 94.5& -& -&-&-& -& - \\
	
 HSE $_{\rm  ICCV{23}}$ \cite{DBLP:conf/iccv/YangSLCCS23}& R50 & 70.6& 80.1& 87.1 & -&89.6& 93.8& 96.0& -&-&-& -& - \\
		PNCA++ $_{\rm  ECCV{20}}$ \cite{DBLP:conf/eccv/TehDT}& R50& 72.2 & 82.0 & 89.2 &93.5& 90.1& 94.5&97.0&98.4&-& -& -&-\\
PLEor $_{\rm  CVPR{23}}$ \cite{DBLP:conf/cvpr/WangCLWO023} & R50& 74.8&  84.5&  91.3& 94.9& 94.4&  96.9&  98.3&  \textbf{98.9} & 86.3&  91.7&  95.1&  96.7\\
 
\cmidrule{1-14} 
 \rowcolor{gray!30}

Our AdvRF & R50& \textbf{76.6} & \textbf{85.3} & \textbf{91.7} &\textbf{95.0}&  \textbf{94.9}& \textbf{97.2}&\textbf{98.6}&\textbf{98.9}&\textbf{88.0}& \textbf{92.5}&\textbf{95.5}&\textbf{96.9}\\
	\toprule[1pt]
	\end{tabular}
    \label{t3}
\end{table*}

\section{Experiments}
\subsection{Experimental Setup}
\textbf{Datasets.} CUB-200-2011 \cite{Branson2014Bird} consists of 200 bird species. We use the first 100 subcategories (5,864 images) for training and the consists of (5,924 images) for testing. 
The Stanford Cars \cite{Krause20133D} includes 196 car models. Similarly, we use the first 98 classes, which contain 8,054 images, for training and the remaining classes, which contain 8,131 images, for testing. Finally, FGVC Aircraft \cite{DBLP:journals/corr/MajiRKBV13} is split into first 50 classes, containing 5,000 images, for training and the remaining 50 classes with 5,000 images, for testing. 
Stanford Online Products (SOP) \cite{DBLP:conf/cvpr/SchroffKP15} is divided into the 11, 318 subcategories (59, 551 images) in training, and the rest 11, 316 classes (60, 502 images) in testing. 
\textit{This split ensures \textbf{no category overlap} between training and testing sets, where all testing categories are strictly unseen during training to evaluate cross-category generalization.}

\noindent\textbf{Implementation Details.}
Our retrieval model is built upon a ResNet-50 backbone~\cite{He2015Deep} initialized with ImageNet pre-trained weights. Input images are resized to 256×256 pixels and randomly cropped to 224×224 during training. We employ Stochastic Gradient Descent with an initial learning rate of $10^{-5}$, weight decay of 0.0001, and momentum of 0.9, using a batch size of 32 distributed across four NVIDIA A100 GPUs. To enhance robustness, standard data augmentations including random cropping, horizontal flipping, and color jittering are applied. The learning rate follows an exponential decay schedule (factor=0.9 every 5 epochs) over 200 training epochs, ensuring stable convergence while mitigating overfitting to category-specific patterns.

\noindent\textbf{Evaluation protocols.} We evaluate the retrieval performance by \textit{Recall@K} with cosine distance, which is average recall scores over all query images in the test set and strictly follows the setting in previous work \cite{DBLP:conf/cvpr/SongXJS16}. Specifically, for each query, our model returns the top $ K $ similar images. In the top $ K $ returning images, the score will be 1 if there exists at least one positive image, and 0 otherwise.

\subsection{Ablation Experiments}
\textbf{Efficacy of various constraints.}
The proposed AdvRF, as described in Sec.~\ref{adv}, is optimized through a combination of four loss functions, each playing a distinct role in guiding AdvRF to capture category-agnostic discrepancies. Tab.~\ref{t1} presents quantitative comparisons across various constraint combinations. Initially, we use ResNet-50 \cite{He2015Deep} with only the classification loss $\mathcal{L}_\mathcal{C}$, achieving 66.3\% Recall@1 accuracy on the CUB-200-2011 dataset. By introducing the reconstruction feedback loss $\mathcal{L}_\mathcal{G}^{RecF}$ and the retrieval feedback loss $\mathcal{L}_\mathcal{R}^{RetF}$, we synergize category-aware discrepancy localization from retrieval models with category-agnostic feature learning from reconstruction models, obtaining a performance of 76.8\%. When removing $\mathcal{L}_\mathcal{G}^A$ or $\mathcal{L}_\mathcal{G}^R$, the reconstruction model’s sensitivity to residual discrepancies from the retrieval model decreases, leading to reduced performance. Additionally, when removing $\mathcal{L}_\mathcal{R}^A$ or $\mathcal{L}_\mathcal{R}^R$, the retrieval model struggles to accurately evaluate the discrepancy localization, which also results in decreased performance. Finally, for efficient deployment, we introduce a category-agnostic discrepancy parameterization loss $\mathcal{L}_{\mathcal{P}}$ to distill category-agnostic discrepancy representations into the retrieval model, enabling real-time retrieval without acceptable accuracy loss.

\noindent\textbf{Different types of feedback.}
Tab.~\ref{t2} presents a comparison of the retrieval performance for retrieving visually similar objects using different types of feedback. Directly using classification-based feedback fundamentally couples discrepancy modeling with predefined category supervision, inadvertently embedding category-specific semantics into the retrieval representations, resulting in a performance of 66.3\%. 
When directly training a reconstruction model and using its encoder outputs for retrieval, the model focuses on modeling the entire image appearance, including irrelevant background information, rather than emphasizing key visual discrepancies, which leads to a lower performance. 
In contrast, our AdvRF effectively combines the advantages of category-aware discrepancy localization from retrieval models with category-agnostic feature learning from reconstruction models to model visual discrepancies using category-agnostic representations, thus achieving a performance of 76.6\%.

\subsection{Comparisons with the State-of-the-Arts}
\textbf{Fine-grained image retrieval.}
Our AdvRF demonstrates superior performance across all three FGIR benchmarks (CUB-200-2011, Stanford Cars-196, FGVC Aircraft), significantly outperforming existing state-of-the-art methods (Tab.~\ref{t3}). 
Localization-based approaches (e.g., CaRA \cite{DBLP:journals/pami/WangCWLOT24}, FRPT \cite{DBLP:conf/aaai/WangCWLO023}) and metric-learning frameworks (e.g., IDML \cite{DBLP:journals/pami/WangZZZL24}, HIST \cite{DBLP:conf/cvpr/LimYP022}) demonstrate effectiveness in capturing fine-grained visual discrepancies. However, their inherent coupling of discrepancy modeling with predefined category supervision embeds category-specific semantics into retrieval representations, fundamentally limiting performance breakthroughs in unseen category generalization. 
Therefore, AdvRF introduces an adversarial reconstruction mechanism that decouples discrepancy modeling from categorical supervision through iterative training between the retrieval and reconstruction models. 
This mechanism explicitly grounds visual discrepancies in appearance cues rather than seen category semantics, thereby achieving significant performance gains in generalization to unseen categories.

\noindent\textbf{Coarse-grained image retrieval.} 
To further validate AdvRF’s generalization capability, we evaluate it on a large-scale coarse-grained benchmark, \textit{ i.e.}, Stanford Online Products, in Tab.~\ref{t4}. 
The framework’s synergy between the precise discrepancy localization of the retrieval model and the category-agnostic representation learning capability of the reconstruction model can represent objects using category-agnostic description. 
Hence, AdvRF not only captures subtle inter-class differences in fine-grained settings but also maintains robustness to coarse-grained semantic gaps, thus obtaining a better performance on SOP.

\begin{table} \centering
\caption{Recall@k for k = 1, 10, 100, 1000 on Stanford Online Products (SOP).}
\begin{tabular}{l||cccc}
\toprule[1pt]

Method & 1 & 10 & 100 & 1000 \\ 
\toprule[0.7pt]

MS \cite{DBLP:conf/cvpr/WangHHDS19} $_{\rm  CVPR{19}}$ & 78.2 & 90.5 & 96.0 & 98.7 \\
NSM \cite{DBLP:conf/bmvc/ZhaiW19} $_{\rm  BMVC{21}}$ & 79.5 & 91.5 & 96.7 & - \\
DCML \cite{DBLP:conf/cvpr/ZhengWL021} $_{\rm  CVPR{21}}$  & 79.8& 90.8 & 95.8 & 95.8 \\
ETLR \cite{DBLP:conf/cvpr/KimKCK21} $_{\rm  CVPR{21}}$ & 79.8& 91.1 & 96.3 & - \\
MRML-PA \cite{DBLP:conf/iccv/ZhengZL021} $_{\rm  ICCV{21}}$ & 79.9 & 90.7 & 96.1 & - \\
HSE \cite{DBLP:conf/iccv/YangSLCCS23} $_{\rm  ICCV{23}}$ &80.0& 91.4& 96.3&-\\
DAS \cite{DBLP:conf/eccv/LiuHZYTW22}  $_{\rm  ECCV{22}}$ & 80.6& 91.8& 96.7& 99.0 \\
CEP \cite{DBLP:conf/eccv/BoudiafRZGPPA20} $_{\rm  ECCV{20}}$& 81.1 & 91.7 & 96.3 & 98.8 \\
PNCA++  \cite{DBLP:conf/eccv/TehDT} $_{\rm  ECCV{20}}$ & 81.4 & 92.4 & 96.9 & 99.0 \\
IBC \cite{DBLP:conf/icml/SeidenschwarzEL21} $_{\rm  ICML{21}}$ & 81.4 & 91.3 & 95.9 & -\\
HIST \cite{DBLP:conf/cvpr/LimYP022} $_{\rm  CVPR{22}}$ &81.4 & 92.0& 96.7& - \\
IDML \cite{DBLP:journals/pami/WangZZZL24} $_{\rm  TPAMI{24}}$&81.5& 92.3& 54.8& 51.3\\
CaRA \cite{DBLP:journals/pami/WangCWLOT24} $_{\rm  TPAMI{24}}$ & 82.4 & 92.6 & 97.0 &99.0\\
\toprule[0.7pt]

 \rowcolor{gray!30}
Our AdvRF & \textbf{84.2} & \textbf{93.7} & \textbf{97.6} &\textbf{99.1}\\
\toprule[1pt]

\end{tabular}
\label{t4}
\end{table}

\subsection{Further Analysis}

\textbf{Investigation on the updating ratio $\delta$.} 
Tab.~\ref{t5} showcases the accuracy of various updating ratios in Eqn.~\ref{eq3}. Notably, as the ratio increases, retrieval performance declines, indicating that excessive updates to the generator cause it to rely too heavily on the current learning parameters, making it harder to fine-tune the parameters within the retrieval model. Conversely, a lower ratio preserves sensitivity to prior knowledge, forcing the generator to rely more on the retrieval model's features, thereby providing more accurate features by modifying the retrieval model's parameters.

\begin{table}[t]
    \centering
    \caption{Evaluation results on CUB-200-2011 of light-weight generator trained with different updating ratio $\delta$ in Eqn.~\ref{eq3}.}
    \begin{tabular}{cccccc}
    \toprule[1pt]
    Ratio $\delta$&0.1&0.2&0.4&0.6&0.8\\
    \toprule[0.7pt]
         R@1&75.4\%&76.6\%&74.8\%&74.4\%&73.9\%\\ 
         \toprule[1pt]
    \end{tabular}
    \label{t5}
\end{table}

\begin{table}\centering
\setlength{\tabcolsep}{7pt}
\caption{Results comparing to various pattern maps based on Recall@K on CUB-200-2011.}
\begin{tabular}{ccccc}
\toprule[1pt]
Method & R@1 & R@2 & R@4 & R@8 \\
\toprule[0.7pt]
CAM \cite{DBLP:journals/ijcv/SelvarajuCDVPB20} & 69.8\% & 79.7\% & 84.2\%  &91.6\% \\
Bounding box & 73.9\% & 82.6\%& 90.5\% & 94.2\% \\
\toprule[0.7pt]

Our AdvRF &\textbf{76.6\%} & \textbf{85.3\%} & \textbf{91.7\%} &\textbf{95.0\%}\\
\toprule[1pt]
\end{tabular}
\label{t6}
\end{table}

\begin{table}
    \centering
    \setlength{\tabcolsep}{9pt}
    \caption{Effect on the reconstruction ability with different reconstructed manners on CUB-200-2011. }
    \begin{tabular}{lll}
    \toprule[1pt]
      Reconstruction Manner   &  Recall@1 & Recall@2 \\
         \toprule[0.7pt]
         Non-adversarial Recon. & 72.6\%& 82.4\% \\
          Adversarial Recon. & \textbf{76.6\%}$_{+4.0}$ & \textbf{85.3\%}$_{+2.9}$\\
         \toprule[1pt]
    \end{tabular}
    \label{t7}
\end{table}

\noindent\textbf{Visual discrepancy localization with different manners.}
Switching the visual discrepancy localization method provides insights into acquiring category-agnostic discrepancies. As Tab.~\ref{t6} indicates, shifting from our discrepancy decoupling strategy to a fixed localization method causes a significant performance drop, nearing the accuracy of fine-tuning a pre-trained model. Specifically, using class activation maps or dataset-provided bounding boxes for localization often leads to imprecise results, including background and missing critical discrepancies. In contrast, our decoupling strategy allows AdvRF to accurately locate visual discrepancies and generate precise category-agnostic representations, consistently improving performance.

\noindent\textbf{Adversarial learning between the retrieval and reconstruction models.}
In AdvRF, the reconstruction model exposes residual discrepancies overlooked by the retrieval model, forcing it to improve localization accuracy, while the refined signals from the retrieval model guide the reconstruction model to enhance its reconstruction capability. When the reconstruction model only uses the discrepancies localized by the retrieval model, rather than its residual discrepancies, it struggles to provide more comprehensive feedback, resulting in a performance of 72.6\% as shown in the first row of Tab.~\ref{t7}. This implicitly indicates that adversarial learning creates a self-reinforcing cycle, where both the retrieval and reconstruction models iteratively challenge and reinforce each other.

\begin{figure}[t]
\begin{center}
   \includegraphics[width=1\linewidth]{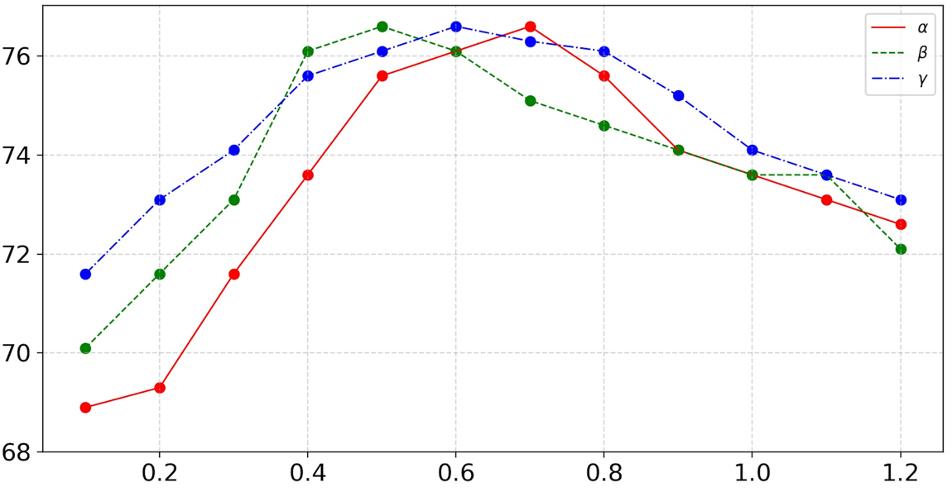}
\end{center}
\vspace{-1em}
\caption{Analyses of hyper-parameters $\alpha$, $\beta$ and $\gamma$ in Eq.~\ref{eq14} and \ref{eq15}. Results denote Recall@1 accuracy on CUB-200-2011.}
\label{loss}
\end{figure}

\noindent{\bf Hyper-parameter analyses.} We conduct sensitivity analyses of the hyperparameters in Eq.~\ref{eq14} and~\ref{eq15}, with evaluation results presented in Fig.~\ref{loss}. The performance of our AdvRF shows slight sensitivity to variations in $\alpha$, $\beta$, $\gamma$, and $\delta$. In our experiments, the default values are set to $\alpha = 0.7$, $\beta = 0.5$, $\gamma = 0.6$, respectively.

\begin{figure}[t]
\begin{center}
   \includegraphics[width=1\linewidth]{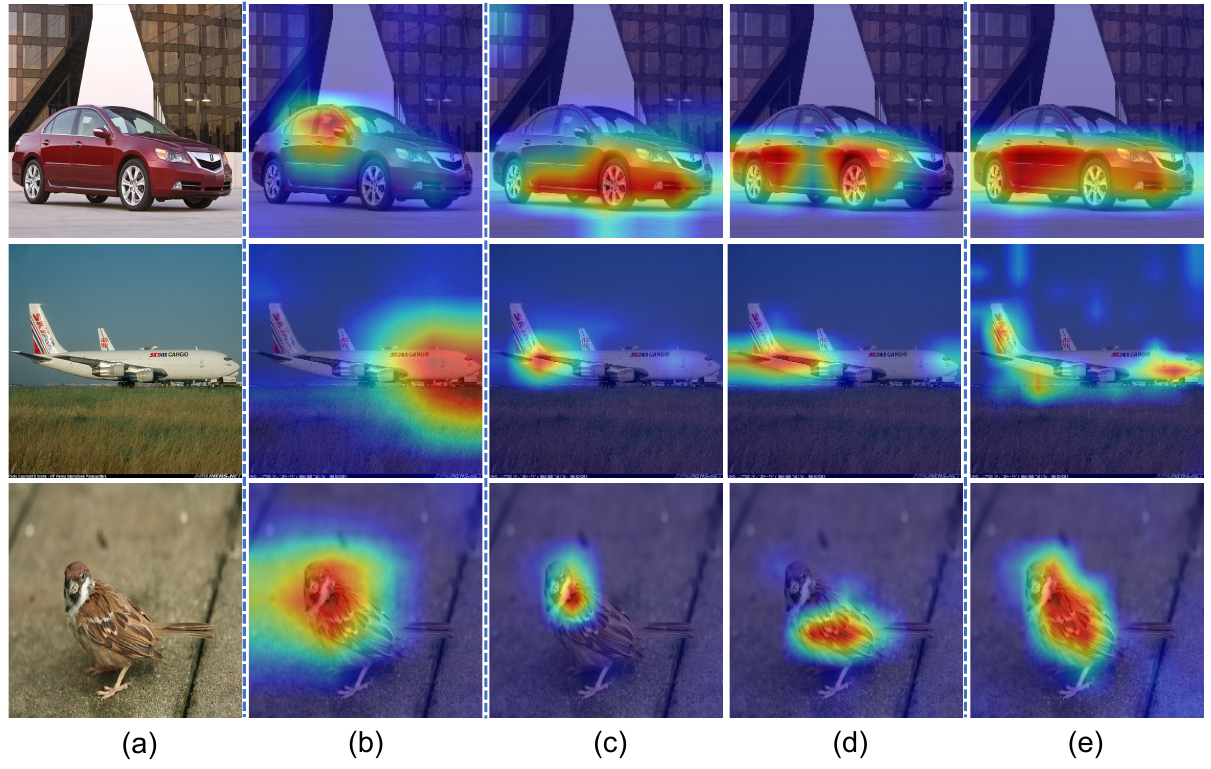}
\end{center}
\setlength{\abovecaptionskip}{-0.2cm} 
   \caption{Visualizations of pattern maps using different feedback: (a) inputs, (b) classification-based feedback, (c) non-discrepancy reconstruction-based feedback, (d) discrepancy reconstruction-based feedback, and (e) our AdvRF.
   }
   \label{activation}
\end{figure}

\noindent{\bf Effect of pattern maps with various feedback.} 
Fig.~\ref{activation} illustrates the impact of various feedback signals on discrepancy localization. When using classification-based feedback, the model struggles to accurately localize discrepancies. However, formulating FGIR as a visual discrepancy reconstruction task enhances discrepancy localization through reconstruction-based feedback. Notably, limited reconstruction signals, such as using only discrepancy-based feedback ($\mathbf{C_A}$ in Eq.~\ref{eq4}) or non-discrepancy-based feedback ($\mathbf{C_R}$ in Eq.~\ref{eq4}), may cause pattern maps to overlook certain discrepancies. Our results suggest that combining comprehensive reconstruction feedback creates a self-reinforcing cycle, where both the retrieval and reconstruction models iteratively challenge and strengthen each other.

\begin{figure}
     \centering
     \begin{subfigure}[b]{\linewidth}
         \begin{minipage}{0.48\linewidth}
	
		\centering
		\includegraphics[width=1\linewidth]{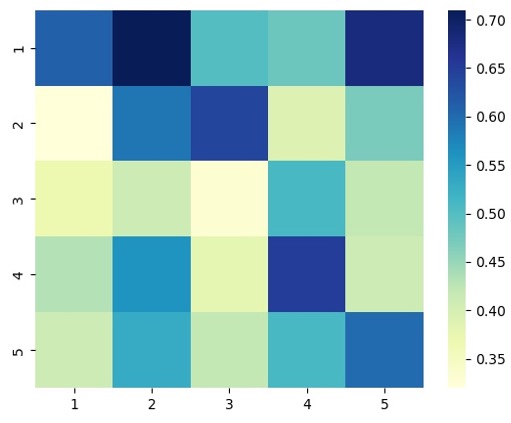}
		\caption{Category-related Embeddings}
		\label{fig:side:a}
	\end{minipage}
	\begin{minipage}{0.48\linewidth}
		\centering
		\includegraphics[width=1\linewidth]{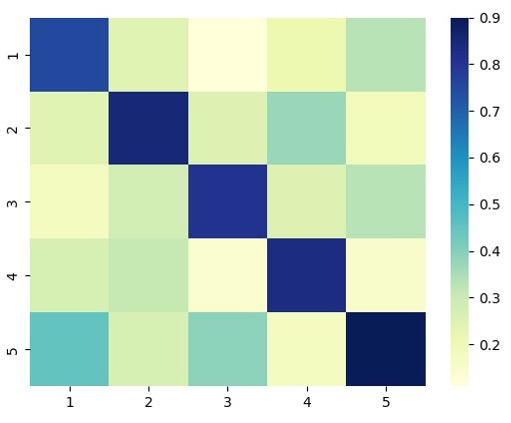}
		\caption{Category-agnostic Embeddings}
		\label{fig:side:b}
	\end{minipage}
     \end{subfigure}
    \caption{Evaluating the efficacy of category-related embeddings from baseline \cite{He2015Deep} versus category-agnostic embeddings from our AdvRF, with their similarity in grid formats.}
        \label{fig:three graphs}
   
\end{figure}

\noindent\textbf{Analysis of category-agnostic discrepancies.}
We employ an indirect method to interpret category-agnostic discrepancies by comparing the similarities between category-agnostic retrieval embeddings produced by our AdvRF and category-related retrieval embeddings generated by the classification-based feedback illustrated in Fig.~\ref{introd1}.
As illustrated in Fig.~\ref{fig:three graphs}, we compute these similarities for the ten high-similarity images across five novel subcategories.
Our analysis reveals that category-agnostic embeddings effectively highlight nearest sample pairs that belong to the same subcategory. In contrast, category-related retrieval embeddings struggle to identify these high-similarity images, as they tend to capture visual discrepancies associated with the semantics of base categories. Overall, our AdvRF successfully learns category-agnostic descriptions for unseen subcategories by formulating FGIR tasks as a visual discrepancy reconstruction process.

\section{Conclusion}
In this paper, we introduce AdvRF, which acquires category-agnostic visual discrepancies by formulating FGIR as a visual discrepancy reconstruction task. 
AdvRF designs an adversarial pipeline: the reconstruction model exposes residual discrepancies overlooked by the retrieval model, forcing it to improve localization accuracy, while the refined signals from the retrieval model guide the reconstruction model to improve its reconstruction ability. 
As a result, AdvRF precisely localizes visual differences and encodes them into category-agnostic representations. 
This representation is then transferred to the retrieval model through knowledge distillation for efficient deployment. 
Importantly, our algorithm is end-to-end trainable and achieves state-of-the-art performance on the widely-used fine-grained and coarse-grained retrieval datasets.

\noindent\textbf{Acknowledgements.}
This work is supported in part by the National Natural Science Foundation of China (NSFC) under Grant (No.61932020),
and the Taishan Scholar Program of Shandong Province (tstp20221128).

{
    \small
    \bibliographystyle{ieeenat_fullname}
    \bibliography{main}
}

\end{document}